# NIMFA : A Python Library for Nonnegative Matrix Factorization

**Marinka Žitnik**  MARINKA.ZITNIK@STUDENT.UNI-LJ.SI
**Blaž Zupan**  BLAZ.ZUPAN@FRI.UNI-LJ.SI
*Faculty of Computer and Information Science*
*University of Ljubljana*
*SI-1000 Ljubljana, Tržaška 25, Slovenia*

## Abstract

NIMFA is an open-source Python library that provides a unified interface to nonnegative matrix factorization algorithms. It includes implementations of state-of-the-art factorization methods, initialization approaches, and quality scoring. It supports both dense and sparse matrix representation. NIMFA's component-based implementation and hierarchical design should help the users to employ already implemented techniques or design and code new strategies for matrix factorization tasks.

**Keywords:** nonnegative matrix factorization, initialization methods, quality measures, scripting, Python

## 1. Introduction

As a method to learn parts-based representation, a nonnegative matrix factorization (NMF) has become a popular approach for gaining new insights about complex latent relationships in high-dimensional data through feature construction, selection and clustering. It has recently been successfully applied to many diverse fields such as image and signal processing, bioinformatics, text mining, speech processing, and analysis of multimedia data (Cichocki et al., 2009). NMF's distinguishing feature is imposition of nonnegativity constraints, where only non-subtractive combinations of vectors in original space are allowed (Lee and Seung, 1999, 2001). Specific knowledge of the problem domain can be modelled by further imposing discriminative constraints, locality preservation, network-regularization or constraint on sparsity.

We have developed a Python-based NMF library called NIMFA which implements a wide variety of useful NMF operations and its components at a granular level. Our aim was both to provide access to already published variants of NMF and ease the innovative use of its components in crafting new algorithms. The library intentionally focuses on nonnegative variant of matrix factorization, and in terms of variety of different approaches compares favourably to several popular matrix factorization packages that are broader in scope (PyMF, (http://code.google.com/p/pymf), NMF package (http://nmf.r-forge.r-project.org), and bioNMF (http://bionmf.cnb.csic.es); see Table 1).

|  | NIMFA | PyMF | NMF | bioNMF |
|---|---|---|---|---|
| Language | Python | Python | R, C++ | PHP, Matlab, C |
| License/Copyright | GPL3 | GPL3 | GPL2+ | license-free |
| Hierarchical factorization models | + | − | (+) | − |
| Sparse format support | + | (+) | − | − |
| Web based client | − | − | − | + |
| Quality measures | + | − | + | + |
| Fitted model and residuals tracking | + | − | + | − |
| Algorithm specific parameters | + | + | + | + |
| Advanced initialization methods | + | − | + | (+) |
| Extensive documentation | + | − | + | + |
| Support for multiple runs | + | − | + | + |
| Visualization | (+) | − | + | + |
| Methods / Shared with NIMFA | 11/11 | 10/3 | 5/4 | 3/3 |

Table 1: Feature comparison of NIMFA and three popular matrix factorization libraries. Symbol + denotes full support, (+) partial support and symbol − no support. Last row reports on a number of different NMF algorithms implemented and a number of these that are shared with NIMFA.

## 2. Supported Factorization Methods and Approaches

In a standard model of NMF (Lee and Seung, 2001), a data matrix $V$ is factorized to $V \equiv WH$ by solving a related optimization problem. Nonnegative matrices $W$ and $H$ are commonly referred to as basis and mixture matrix, respectively. NIMFA implements an originally proposed optimization (Lee and Seung, 2001; Brunet et al., 2004) with Euclidean or Kullback-Leibler cost function, along with Frobenius, divergence or connectivity costs. It also supports alternative optimization algorithms including Bayesian NMF Gibbs sampler (Schmidt et al., 2009), iterated conditional modes NMF (Schmidt et al., 2009), probabilistic NMF (Laurberg et al., 2008) and alternating least squares NMF using projected gradient method for subproblems (Lin, 2007). Sparse matrix factorization is provided either through probabilistic (Dueck and Frey, 2004) or alternating nonnegativity-constrained least squares factorization (Kim and Park, 2007). Fisher local factorization (Wang et al., 2004; Li et al., 2001) may be used when dependency of a new feature is constrained to a given small number of original features. Crisp relations can be revealed by binary NMF (Zhang et al., 2007).

NIMFA also implements several non-standard models. These comprise nonsmooth factorization $V \equiv W S(\theta) H$ (Pascual-Montano et al., 2006) and multiple model factorization for simultaneous treatment of several input matrices and their factorization with the same basis matrix $W$ (Zhang et al., 2011).

All mentioned optimizations are incremental and start with initial approximation of matrices $W$ and $H$. Appropriate choice of initialization can greatly speed-up the convergence and increase the overall quality of the factorization results. NIMFA contains implementations of popular initialization methods such as nonnegative double singular value decomposition (Boutsidis and Gallopoulos, 2007), random C and random Vcol algorithms (Albright et al., 2006). User can also completely specify initial factorization by passing fixed factors or choose any inexpensive method of randomly populated factors.

Factorization rank, choice of optimization method, and method-specific parameters jointly define the quality of approximation of input matrix *V* with the factorized system. NIMFA provides a number of quality measures ranging from standard ones (e.g., Euclidean distance, Kullback-Leibler divergence, and sparseness) to those more specific like feature scoring representing specificity to basis vectors (Kim and Park, 2007).

## 3. Design and Implementation

NIMFA has hierarchical, modular, and scalable structure which allows uniform treatment of numerous factorization models, their corresponding factorization algorithms and initialization methods. The library enables easy integration into user's code and arbitrary combinations of its factorization algorithms and their components. NIMFA's modules encompass implementations of factorization (`nimfa.methods.factorization`) and initialization algorithms (`nimfa.methods.seeding`), supporting models for factorization, fitted results, tracking and computation of quality and performance measures (`nimfa.models`), and linear algebra helper routines for sparse and dense matrices (`nimfa.utils`).

The library provides access to a set of standard data sets (`nimfa.datasets`), including those from text mining, image processing, bioinformatics, functional genomics, and collaborative filtering. Module `nimfa.examples` stores scripts that demonstrate factorization-based analysis of these data sets and provide examples for various analytic approaches like factorization of sparse matrices, multiple factorization runs, and others.

The guiding principle of constructing NIMFA was a component-oriented architecture. Every block of the algorithms, like data preprocessing, initialization of matrix factors, overall optimization, stopping criteria and quality scoring may be selected from the library or defined in a user-script, thus seamlessly enabling experimentation and construction of new approaches. Optimization process may be monitored, tracking residuals across iterations or tracking fitted factorization model.

NIMFA uses a popular Python matrix computation package `NumPy` for data management and representation. A drawback of the library is that is holds matrix factors and fitted model in main memory, raising an issue with very large data sets. To address this, NIMFA fully supports computations with sparse matrices as implemented in `SciPy`.

## 4. An Example Script

The sample script below demonstrates factorization of medulloblastoma gene expression data using alternating least squares NMF with projected gradient method for subproblems (Lin, 2007) and Random Vcol (Albright et al., 2006) initialization algorithm. An object returned by `nimfa.mf_run` is fitted factorization model through which user can access matrix factors and estimate quality measures.

```python
import nimfa
V = nimfa.examples.medulloblastoma.read(normalize = True)
fctr = nimfa.mf(V, seed='random_vcol', method='lsnmf', rank=40, max_iter=65)
fctr_res = nimfa.mf_run(fctr)

print 'Rss:_%5.4f,_Evar:_%5.4f' % (fctr_res.fit.rss(), fctr_res.fit.evar())
```

```
print 'K-L_divergence:_%5.4f' % fctr_res.distance(metric = 'kl')
print 'Sparseness,_W:_%5.4f,_H:_%5.4f' % fctr_res.fit.sparseness()
```

Running this script produces the following output, where slight differences in reported scores across different runs can be attributed to randomness of the Random Vcol initialization method.

```
Rss: 0.1895, Evar: 0.9998
K-L divergence: 38.6581
Sparseness, W: 0.7279, H: 0.8739
```

## 5. Availability and Requirements

NIMFA is a Python-based package requiring `SciPy` version 0.9.0 or higher. It is available under the GNU General Public License (GPL) version 3. The latest version with documentation and working examples can be found at `http://nimfa.biolab.si`.

## Acknowledgments

We would like to acknowledge support for this project from the Google Summer of Code 2011 program and from the Slovenian Research Agency grants P2-0209, J2-9699, and L2-1112.